\documentclass[letterpaper]{article} % DO NOT CHANGE THIS
\usepackage{aaai23}  % DO NOT CHANGE THIS
\usepackage{times}  % DO NOT CHANGE THIS
\usepackage{helvet}  % DO NOT CHANGE THIS
\usepackage{courier}  % DO NOT CHANGE THIS
\usepackage[hyphens]{url}  % DO NOT CHANGE THIS
\usepackage{graphicx} % DO NOT CHANGE THIS
\usepackage{natbib}  % DO NOT CHANGE THIS AND DO NOT ADD ANY OPTIONS TO IT
\usepackage{caption} % DO NOT CHANGE THIS AND DO NOT ADD ANY OPTIONS TO IT
\usepackage{algorithm}
\usepackage{algorithmic}

\usepackage{multirow}
\usepackage{booktabs} % toprule表格画横线
\usepackage{bbding}  % 对号叉号

\usepackage{amsthm,amsmath,amssymb}  % 花体
\usepackage{mathrsfs}
\usepackage{makecell} % 表格换行

\title{CoMAE: Single Model Hybrid Pre-training on Small-Scale RGB-D Datasets}

\author {
    Jiange Yang,\textsuperscript{\rm 1}
    Sheng Guo, \textsuperscript{\rm 2}
    Gangshan Wu, \textsuperscript{\rm 1}
    Limin Wang \textsuperscript{\rm 1}\thanks{Corresponding author.}
}

\affiliations {
    \textsuperscript{\rm 1} State Key Laboratory for Novel Software Technology, Nanjing University, China \quad
    \textsuperscript{\rm 2} MYbank, Ant Group, China\\
    \{jiangeyang.jgy, guosheng1001\}@gmail.com, \{gswu,lmwang\}@nju.edu.cn
}

\usepackage{bibentry}
\begin{document}

\maketitle

\begin{abstract}
% However, human annotation of extra large-scale RGB datasets are generally time-consuming. 
Current RGB-D scene recognition approaches often train two standalone backbones for RGB and depth modalities with the same Places or ImageNet pre-training.
However, the pre-trained depth network is still biased by RGB-based models which may result in a suboptimal solution. In this paper, we present a single-model self-supervised hybrid pre-training framework for RGB and depth modalities, termed as {\bf CoMAE}. Our CoMAE presents a curriculum learning strategy to unify the two popular self-supervised representation learning algorithms: contrastive learning and masked image modeling. 
Specifically, we first build a patch-level alignment task to pre-train a single encoder shared by two modalities via cross-modal contrastive learning. Then, the pre-trained contrastive encoder is passed to a multi-modal masked autoencoder to capture the finer context features from a generative perspective. 
In addition, our single-model design without requirement of fusion module is very flexible and robust to generalize to unimodal scenario in both training and testing phases. 
Extensive experiments on SUN RGB-D and NYUDv2 datasets demonstrate the effectiveness of our CoMAE for RGB and depth representation learning. 
In addition, our experiment results reveal that CoMAE is a data-efficient representation learner. Although we only use the small-scale and unlabeled training set for pre-training, our CoMAE pre-trained models are still competitive to the state-of-the-art methods with extra large-scale and supervised RGB dataset pre-training. Code will be released at https://github.com/MCG-NJU/CoMAE. 

%AAAI要求不让用超链接，代码先直接放普通字体
%\href{https://github.com/MCG-NJU/CoMAE}{https://github.com/MCG-NJU/CoMAE}.

\end{abstract}

\section{Introduction}

	\begin{figure}[htb]
		\centering
		\includegraphics[width=0.45\textwidth]{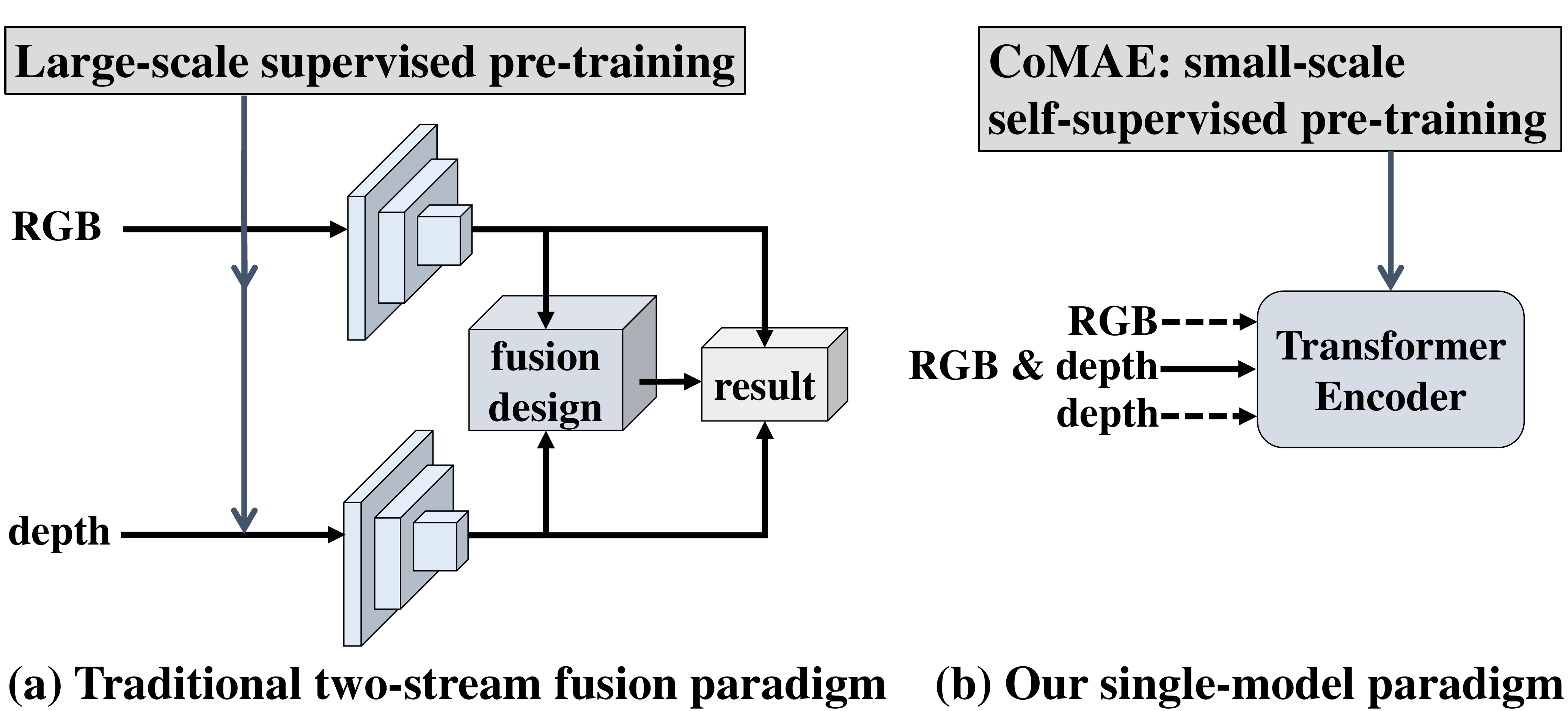} %  [width=0.9\columnwidth] Reduce the figure size so that it is slightly narrower than the column. Don't use precise values for figure width.This setup will avoid overfull boxes.
		\caption{Comparison of RGB-D scene recognition pipeline. (a) Traditional methods often use supervised pre-training to initialize CNNs and train two separate networks for RGB and depth modalities. Then, these two networks are combined with a fusion module to yield final recognition results. (b) Our CoMAE presents {\bf single-model} and {\bf self-supervised} pre-training for RGB and depth two modalities with a Transformer encoder. Our pre-trained singel-model encoder could be deployed for unimodal or multimodal input due its flexibility of self-attention operation.}
		\label{fig:motivation}
	\end{figure}

Thanks to the flourishing development of deep learning, a series of vision tasks, including image classification \cite{he2016deep}, object detection \cite{ren2015faster}, and semantic segmentation \cite{chen2018encoder}, have achieved impressive success by learning discriminative features in an end-to-end manner.
Scene recognition~\cite{zhou2014learning,WangGHX017} is a more challenging task that could benefit from these object-centric understanding. However, it also tends to depend on more representative features than object-centric tasks due to illumination changes, complex spatial layout, and the diversity of objects and their correlations. Therefore, with the widespread use of cost-affordable depth sensors, e.g., Kinect, the research community has witnessed the advancement of RGB-D scene recognition~\cite{silberman2012indoor,song2015sun}. 
The depth modality can provide the cues on the shape of a local geometry and is less sensitiveness to illumination changes and texture variation compared with RGB in this task. 
%Therefore, RGB-D scene recognition \cite{wang2016modality, song2018learning} has attract more and more researchers in the last several years.
%最后一句可能逻辑不太通顺
	
As shown in Fig.~\ref{fig:motivation}, the most straightforward method for RGB-D scene recognition~\cite{zhu2016discriminative,xiong2021ask,du2021cross} is to train two separate networks (CNNs) for two modalities, and combine their results with some fusion modules to yield the final recognition results. In this paradigm, depth data is expected to learn fine and complementary geometrical features to enhance the performance of RGB CNN. However, data scarcity is an important issue for depth modality and training from scratch is prone to over-fitting. Therefore, some methods \cite{zhu2016discriminative} try to use RGB pre-trained models on ImageNet~\cite{deng2009imagenet} or Places~\cite{zhou2014learning}  to initialize the depth CNNs to mitigate this issue. Yet, the modality gap between RGB and depth makes this cross-modal pre-training strategy to be biased and less effective~\cite{song2017depth,du2021cross}. 
Meanwhile, this recognition paradigm of fusing two-stream CNNs makes the whole scene recognition pipeline less efficient and incapable of handling missing modalities in practice. 
Therefore, it still remains as a challenge that {\em how to effectively learn RGB and depth representations on the relatively small-scale RGB-D datasets without using extra pre-training or data, that could be easily and efficiently generalized to downstream tasks with multimodal or unimodal data.}

Recently Transformer~\cite{transformer} has shown the unique advantages in multi-modal understanding tasks, such as image-text retrieval~\cite{radford2021learning}, image-text caption~\cite{coca22}, and video-audio understanding~\cite{vatt21}. The attention operation exhibits high flexibility in handling the multi-modality inputs, and can yield a unified architecture to process different modalities with a single encoder~\cite{polyvit,girdhar2022omnivore}. However, these multi-modal transformers are all pre-trained on large-scale datasets of millions of samples with human labels or noisy texts. It still remains unclear how to develop an effective self-supervised multi-modal transformer on the relatively small-scale RGB-D datasets. Inspired by the results in VideoMAE~\cite{tong2022videomae}, we aim to adapt the data-efficient masked image autoencoder (MAE)~\cite{he2022masked} to multi-modal and small-scale RGB-D datasets for better representation learning.

Based on the above analysis, in this paper, we present a single-model hybrid pre-training method on the relatively small-scale RGB-D datasets. Our method is based on a self-supervised multi-modal transformer, termed as {\bf CoMAE}. To better learn the multi-modal information from limited training samples, we devise a self-supervised curriculum learning framework. Our CoMAE unifies the two types of self-supervised learning paradigms (contrastive learning and masked image modeling), and trains a single encoder for both RGB and depth modalities. 
Specifically, we first build a cross-modal patch-level alignment task to guide the encoder training in a self-supervised manner. This contrastive learning objective would encourage the encoder to capture the coarse structure information to build correspondence between modalities. Then, the learned encoder will be passed into a multi-modal masked autoencoder to perform another round of pre-training from a generative perspective. The multi-modal generative objective would pose a more challenging pre-training task that requires for more detailed granularity to predict the exact pixel values of masked tokens. Unlike the original MAE~\cite{he2022masked}, our CoMAE presents a shared encoder and decoder among RGB and depth modalities and acts as a kind of regularizer to guide pre-training.
In addition, the pre-trained transformer encoder pre-trained by our CoMAE breaks the conventional two-stream fusion paradigm and can handle both multi-modal and unimodal inputs in the subsequent downstream deployment. In summary, our main contribution is threefold:
	\begin{itemize}
	    %第一个在小数据集上探索RGB-D表示学习或场景识别的自监督方法，两个pretext task。
		\item We present a single-model self-supervised hybrid pre-training framework (CoMAE) for RGB-D representation learning with application on scene recognition. Our CoMAE presents a minimalist design in curriculum learning manner to unify two types of popular self-supervised learning methods for RGB-D pre-training on a relatively small-scale dataset.
		% which consists of the cross-modal patch-level contrast and multi-modal masked reconstruction. To the best of our knowledge, it is the first attempt to explore self-supervised representation learning for RGB-D scene recognition on very small-scale RGB-D datasets.
		
		%参数共享节约显存 单模态移植起来也简单，finetune受益于单模态数据，pre-train好好调一下也是，测试的时候只有单模态数据也可以推理,omnivore和multimae的优点吧。这种非常naive的融合方式也是有优点的。
		\item Our proposed single-model encoder is modality-agnostic, which simultaneously deal with RGB and depth with a parameter-sharing encoder. Hence it is flexible to generalize to various settings of multi-modal or unimodal data in both training and testing phases.
		
		%尽管是数据饥饿的transformer模型，小数据集预训练表现可以，文中要体现大数据集的扩展潜力（但是大数据分布不能差异太大），同时强调这是初步探索。
		\item For first time, our CoMAE demonstrates that a multi-modal transformer could be successfully trained on limited RGB-D samples, without using extra data or pre-trained models. Our CoMAE pre-trained models are competitive to those methods with strong and large-scale data pre-training on the two challenging RGB-D scene recognition datasets of NYUDv2 and SUN RGB-D.
	\end{itemize}

\section{Related Work}

\textbf{RGB-D Scene Recognition.}
 Several previous works explicitly modeled the object correlations to obtain a more comprehensive understanding of the scene, such as \cite{yuan2019acm} and \cite{song2019image}. Recently some works focused on excavating the inter-modality and intra-modality relationships. For example, MSN \cite{xiong2020msn} simultaneously extracted the local modal-consistent and global modal-specific features. \cite{du2021cross} proposed to enhance the modality-specific discriminative ability by a cross-modal translation fashion, which both effectively transfers complementary cues to the backbone network. However, these methods all use RGB pre-trained models on ImageNet or Places to initialize the depth CNNs, which is biased and less effective. In addition, the two-stream fusion paradigm makes the recognition less efficient and incapable of handling missing modalities in practice. Instead, our CoMAE does not rely any extra data or pre-trained models and can handling both inputs of multi-modal or unimodal data.

% However, these methods all develop dual backbones which both heavily rely on the same Places or ImageNet pre-training. In addition, we believe that our modality-agnostic single model design is more flexible and general than two-stream manner with modality-specific models as well as performs more robust when only single modality data is available.

\subsubsection{Self-Supervised Learning.}
    Self-supervised representation learning automatically constructs pretext tasks from data itself to learn rich representations that can be transferred to downstream tasks. Currently two types of popular pretext tasks have been designed and applied for self-supervised learning: masked image modeling \cite{bao2021beit,he2022masked} and contrastive learning \cite{chen2020simple,he2020momentum,BYOL}. These two types of self-supervised learning methods focus on modeling data distribution from generative and discriminative perspectives. Some methods focuses on improving downstream tasks for dense prediction~\cite{densecl,convmae}. Recently some works \cite{Siamese,denoise} also attempted to unify these two types self-supervised representation learning algorithms through the patch-level contrastive loss between masked prediction tokens and the augmented view from a momentum encoder. Different from these methods, we propose a curriculum learning strategy to combine these two types of self-supervised learning methods in a simpler way. Besides, we extend this setting to multi-modal data and only perform pre-training on very small-scale dataset.
    
        \begin{figure*}[t]
		\centering
		\includegraphics[width=0.97\textwidth]{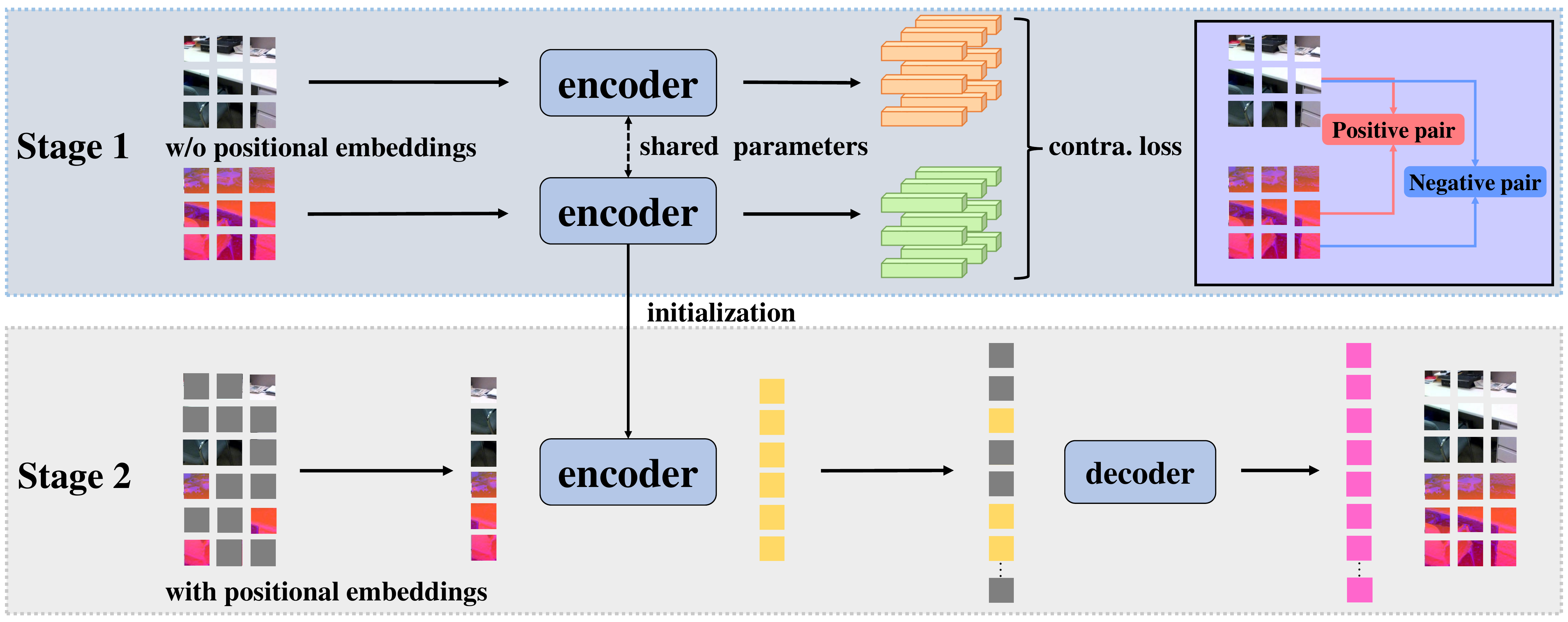} % Reduce the figure size so that it is slightly narrower than the column.
		\label{method}
 		\caption{The CoMAE framework. Our CoMAE performs cross-modal patch-level contrastive pre-training in the first stage (Upon) which provides weights initialization for following reconstruction task and multi-modal masked pre-training in the second stage (Down) which provides weights initialization for downstream recognition task. In the figure, all encoders have the same structure, and the cuboids denote vision tokens of RGB and depth.
   }
		\label{pipelines}
	\end{figure*}
	
\subsubsection{Multi-Modal Transformer.}
    Transformer architectures are very expert in handling multiple input modalities data due its flexibility of self-attention operation. We note that transformers have been used to process multiple modalities in a single encoder such as vision and language \cite{kim2021vilt}, vision and point cloud \cite{wang2022bridged}, as well as video and audio \cite{Bottlenecks}. Recently pre-training multi-modal transformers by masked modeling on large-scale data has become a popular approach towards solving kinds of downstream tasks. \cite{omnimae} jointly pre-trains a single unified model on images and videos on unpaired data with masked modeling. \cite{M3AE} and \cite{bachmann2022multimae} build the multi-modal masked autoencoders for paired image-text data and RGB-depth data, respectively. Significantly different from \cite{bachmann2022multimae}, our CoMAE proposes to use a single decoder for different modalities, while \cite{bachmann2022multimae} used different decoders for different modalities. Meanwhile, our CoMAE conducts multi-modal masked reconstruction on small-scale RGB-D datasets instead of large-scale ImageNet with pseudo depth from depth estimation network. 

\section{Methodology}
    In this section, we describe our CoMAE in detail. First, we give an overview of the CoMAE framework, which is composed of the cross-modal patch-level contrastive pre-training and the multi-modal masked pre-training. Then, we give a technical description on these two components. Finally, we describe how to fine-tune on downstream scene recognition task.
	
	\subsection{Overiew of CoMAE}
% 	Instead of previous approaches which develop two standalone backboness for RGB and depth both obtaining the same hierarchical initialization representations through Places or ImageNet pre-training, we propose a unified single model self-supervised hybrid pre-training framework called CoMAE to simultaneously operate RGB and depth without the need of any sophisticated fusion design. 
As analyzed above, RGB-D scene recognition often uses a two-stream CNN with extra pre-training on large-scale datasets. To handle the issue of representation learning from limited RGB-D samples, we present the CoMAE framework as shown in Fig.~\ref{pipelines}. Our CoMAE aims to train a single unified model for RGB and depth modalities on small-scale datasets with a simple Vision Transformer (ViT)~\cite{dosovitskiy2020image}. We employ a hybrid self-supervised pre-training strategy to progressively tune the parameters of an encoder in curriculum learning manner. First, we propose a relatively easier cross-modal contrastive learning framework to capture the correspondence between two modalities in patch level. This correspondence task will guide transformer encoder to capture structure information to discriminate different patch instances. Then, we present the multi-modal masked autoencoder with a simple parameter sharing scheme in both encoder and decoder among two modalities. This masking and reconstruction task will encourage the transformer encoder to extract detailed granularity to regress the exact pixel values of missing tokens. Meanwhile, the shared encoder-decoder approach would greatly reduce the parameter numbers and also act as a kind of regularizer for pre-training. This customized hybrid training strategy generally follows the principle of curriculum learning~\cite{cl09} and is able to improve the generalization ability of learned representation to the downstream tasks. Thanks to this single-model pre-training scheme, our ConMAE pre-trained encoder could be deployed for fine-tuning with either multi-modal or unimodal inputs, which is efficient to handle the issue of lacking modality in practice.

% The overall framework of the proposed CoMAE is shown in Fig. \ref{pipelines}, which conducts hybrid pre-training in a two-stage manner. In the first stage, we build a patch-level alignment task to pre-train a single encoder shared by two modalities with cross-modal contrastive learning, the trained encoder is instantiated by ViT with random initialization. This pretext task could significantly facilitate further multi-modal integration thanks to its inter-modal alignment ability. In the second stage, we load the parameters of the contrastive encoder pre-trained in the first stage and implement reconstruction in a multi-modal masked autoencoder. This is a challenging self-supervisory task that requires holistic understanding in high-level both for RGB and depth to capture finer features. It is worth nothing that whether cross-modal contrast or multi-modal reconstruction, they enforce the encoder to perceive the spatial layout and the relative position relationship between objects, which plays a more important role on the task of scene recognition than object-centric classification task.
	
\subsection{Cross-modal Patch-level Contrastive Pre-training}
	Inspired by the success of contrastive pre-training in vision and language \cite{li2020learning,MiechASLSZ20,radford2021learning,JiaYXCPPLSLD21}, we present our cross-modal patch-level contrastive (CPC) pre-training. To the best of our knowledge, the joint learning objective of enforcing RGB-depth alignment has not been well explored. Specifically, for each paired RGB-depth image in a training iteration, the input image $r$ and corresponding depth $d$ are separately fed into respective patch projection layers and the shared encoder to generate the individual patch feature representations  $\{f(r)_i|_{i=1}^N\}$ and $\{f(d)_i|_{i=1}^N\}$,  where $i$ is the patch index, $N$ is the total number of patches within an RGB or depth (e.g., $N=196$ for a $224\times 224$ image with a patch size of $16\times 16$). Note that we remove the sine-cosine positional embeddings of ViT to avoid information leakage in contrastive learning. Moreover, different from conventional contrastive learning methods like \cite{he2020momentum, chen2020simple}, our query and dictionary keys are cross-modal patch within a paired RGB-depth. 
	
	Specifically, we use the InfoNCE loss~\cite{oord2018representation} to train our transformer encoder, with a fixed temperature of 0.07. This loss maximizes the similarity of an RGB patch and its corresponding depth patch in the same location, while minimizes similarity to all other patches in corresponding depth map. Our cross-modal contrastive loss function $Loss_{cpc}$ is as follows:
	
% 	 \begin{equation}
%     \footnotesize
%     \label{eq:c()}
%      L_{rgb}(i) = -\log \frac{\exp(s(f(r)_i, f(d)_i)/\tau)}{ \sum\limits_{\substack{k=1}}^{N} \exp(s(f(r)_i, f(d)_k)/\tau)}
%     \end{equation}
    
%     \begin{equation}
%     \footnotesize
%     \label{eq:c()}
%      L_{hha}(i) = -\log \frac{\exp(s(f(d)_i, f(r)_i)/\tau)}{ \sum\limits_{\substack{k=1}}^{N} \exp(s(f(d)_i, f(r)_k)/\tau)}
%     \end{equation}
    
    \begin{equation}
    \footnotesize
    \label{eq:c(1)}
     \ell_{rgb}(i) = -\log \left ( \frac{\exp(s(f(r)_i, f(d)_i)/\tau)}{\sum_{k=1}^{N} \left ( \exp(s(f(r)_i, f(d)_k)/\tau) \right ) }  \right ), 
    \end{equation}
    
    \begin{equation}
    \footnotesize
    \label{eq:c(2)}
    \ell_{depth}(i) = -\log \left ( \frac{\exp(s(f(d)_i, f(r)_i)/\tau)}{\sum_{k=1}^{N} \left ( \exp(s(f(d)_i, f(r)_k)/\tau) \right ) }  \right ), 
    \end{equation}
    
     \begin{equation}
    \footnotesize
    \label{eq:c(3)}
    Loss_{cpc} = \frac{1}{2N} \left (\sum_{i=0}^{N}  \left (  \ell_{rgb}(i)+ \ell_{depth}(i)\right )  \right ).
    \end{equation}
    where $\tau$ and $s(.)$ are temperature coefficient and similarity functions respectively.
    
	This patch-level contrastive pre-training explicitly utilize the inter-modal alignment relations as the self-supervisory signal. It expects the cross-modal alignment may be implicitly learned from paired small-scale RGB-depth datasets. With the above optimized objectives, the encoder network is encouraged to better capture the discriminative local context representations across RGB and depth modalities, which could also facilitate their the following multi-modal masking and reconstruction task. 
% 	Futhermore, the relative position relationship of object to object plays an important role on scene understanding, intuitively our patch-level contrast pre-training also benefits this.
	
%	Of course, we have to admit that the cross-modal patch-level contrast loss between RGB and depth may carry some noisy negative examples since adjacent patches are semantically redundant, thereby there should have further refinement space for negative examples mining.

	\subsection{ Multi-modal Masked Pre-training}

	After the cross-modal contrastive learning, our transformer encoder can learn some structure information to discriminate the corresponding RGB and depth patch pairs. To further enhance its representation power, we propose to perform another round training of transformer encoder in a generative view. Driven by the good self-supervised learning performance of MAE \cite{he2022masked} in images, we design a multi-modal masked autoencoder (MM-MAE) in RGB-D data. RGB and depth are homogeneous modalities and therefore our MM-MAE could easily deal with them with a single encoder and decoder with parameter sharing for both modalities. Our MM-MAE just requires the modality-specific patch projection layers and prediction heads to distinguish the different modalities and reconstruction tasks. This {\em parameter sharing scheme} would greatly reduce the pre-training parameter numbers and also act as a kind of regularizer to improve the pre-training performance.
	
	Specifically, we use the corresponding patch projection layers to obtain the RGB and depth tokens, and then add the fixed and shared 2D sine-cosine positional embeddings for two modalities. All patch tokens of two modalities are concatenated into a sequence. Afterward we randomly mask a subset of RGB and depth tokens both with a $75\%$ masking ratio, and the encoder takes all visible (unmasked) tokens as input. After encoder, our lightweight decoder takes all encoded visible tokens and learnable mask tokens with 2D positional embeddings as input. The mask tokens are shared among two modalities. Finally, we apply two modality-specific prediction heads to reconstruct RGB and depth patches, respectively. Our reconstruction target $\hat{P}_{r}$ and $\hat{P}_{d}$ are per-patch normalization pixels. The whole MM-MAE pipeline can be formally described as follows:
	\begin{equation}
    \footnotesize
    \label{eq:c(4)}
     F = \mathrm{Decoder}(\mathrm{Encoder}(T_{vis\_r}, T_{vis\_d}),T_{mask}), 
    \end{equation}
    \begin{equation}
    \footnotesize
    \label{eq:c(5)}
         \begin{cases}
    \hat{P}_{r}= \mathrm{Linear}_{RGB}(F_{r})
     \\\hat{P}_{d}= \mathrm{Linear}_{depth}(F_{d})
    \end{cases}, 
    \end{equation}
    where $T_{vis_r}$ and $T_{vis_r}$ are visual tokens, $T_{mask}$ is mask token, $\hat{P}$ is the reconstructed patches for RGB or depth modality. We compute the MSE loss between the reconstructed and original pixels of all masked patches. So our MM-MAE loss function $Loss_{mm-mae}$ is as follows:
    \begin{equation}
    \footnotesize
    \label{eq:c(6)}
    Loss_{mm-mae} = \mathrm{MSE}(\hat{P}_{m\_r} , P_{m\_r} ) + \mathrm{MSE}(\hat{P}_{m\_d}, P_{m\_d}).
    \end{equation}
    By reconstructing all the masked RGB and depth tokens, our MM-MAE could enforce the encoder to capture modality-complementary features in a joint space manner while simultaneously preserving modality-specific features.
    
    % \begin{equation}
    % \footnotesize
    % \label{eq:c(5)}
    %  P_{depth} = Lin_{depth}(D(E(T_{vis\_r}, T_{vis\_d}),T_{mask})), 
    %P_{RGB}= Linear_{RGB}(F_{RGB})
    %P_{depth}= Linear_{depth}(F_{depth})
    % \end{equation}
	
	       \begin{figure}[t]
		\centering
		\includegraphics[width=0.435\textwidth]{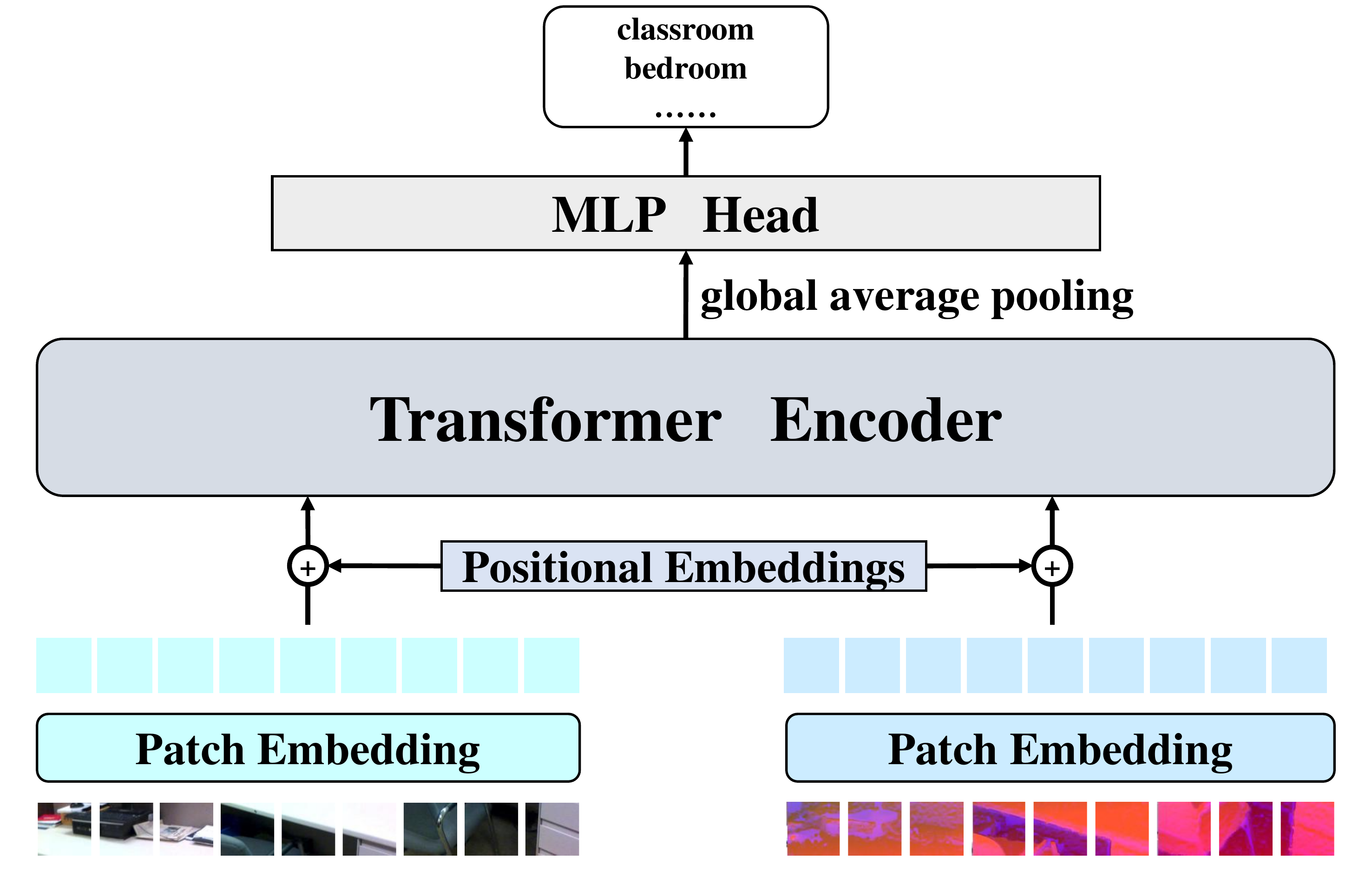} %  [width=0.9\columnwidth] Reduce the figure size so that it is slightly narrower than the column. Don't use precise values for figure width.This setup will avoid overfull boxes.
		\caption{Illustration of the fine-tuning on downstream scene recognition task.}
		%\vspace{-1mm}
		\label{finetune2}
	\end{figure}
	
    \subsection{Fine-tuning on Scene Recognition }
    
    After our CoMAE pre-training, the pre-trained encoder can be directly fine-tuned on the downstream scene recognition task by appending light linear classifier on top. Specifically, we also concatenate all RGB and depth tokens in the same process with the multi-modal masked reconstruction as shown in Fig~\ref{finetune2}. We obtain the global representations via global average-pooling (GAP) instead of a class token. It is worth noting that during fine-tuning process, we randomly drop a modality with a probability of 0.5 as data augmentation to obtain better results. 
    % This suggests that our single model could benefit from unpaired unimodal data during fine-tuning and thus has enormous advancement potential when there are a great quantity of labeled unpaired unimodal data.

	\section{Experiments}
	In this section, we present the experiment results of our CoMAE framework on the task of RGB-D scene recognition. First, we introduce the evaluation datasets and the implementation details of our proposed approach. Then we elaborate the ablation studies on our CoMAE design. We also show some visualization results to further analyze our the effectiveness of our CoMAE. Finally, we evaluate the performance of our CoMAE and compare with the previous state-of-the-art methods. We quantitatively report the average accuracy over all samples (${acc}_s$) and the average accuracy over all scene classes (${acc}_c$) according to the previous evaluation scheme.

	\subsection{Datasets}
	
		\subsubsection{SUN RGB-D Dataset} is the most popular RGB-D scene recognition dataset. It contains RGB-D images from NYU depth v2, Berkeley B3DO \cite{janoch2013category}, and SUN3D \cite{xiao2013sun3d}, which is composed of 3,784 Microsoft Kinect v2 images, 3,389 Asus Xtion images, 2,003 Microsoft Kinect v1 images and 1,159 Intel RealSense images. Following the official setting in \cite{song2015sun}, we only use the images from 19 major scene categories, containing 4,845 / 4,659 train / test images.
	
		\subsubsection{NYU Depth Dataset V2 (NYUDv2)} contains 1449 well labeled RGB-D images for scene recognition. 795 images are for training and 654 are for testing. Following the standard split in \cite{silberman2012indoor}, the 27 scene categories are grouped into 9 major categories and an other category.

	\subsection{Implementation Details} % 有时间需要更改一下

	 We employ the three-channel HHA \cite{gupta2014learning} to encode depth. HHA offers a representation of geometric properties at each pixel, including the horizontal disparity, the height above ground, and the angle with gravity, which has been proven to have better performance to capture the scenes structural and geometrical information than raw depth. Our proposed CoMAE is implemented on the Pytorch toolbox, and we pre-train and fine-tune our models on eight TITAN Xp GPUs using AdamW \cite{loshchilov2017decoupled} optimizer with a weight decay 0.05. In hybrid pre-training stage, we separately train cross-modal patch-level contrastive learning for 75 epochs and multi-modal masked autoencoder for 1200 epochs. The base learning rate is set to $1.0\times10^{-3}$ for pre-training and $5.0\times10^{-4}$ for fine-tuning, as well as the batch size is set to 256 for pre-training and 768 for fine-tuning, the
     effective learning rate follows the linear scaling rule in \cite{goyal2017accurate}: $lr = base\_lr \times batchsize \div 256$. The warm-up and layer decay strategies are used to adjust the learning rate. Finally, it is worth noting that our pre-training uses cropping and flipping as data augmentation following MAE, but our fine-tuning only adds mixup, cutmix and erasure, which is much easier to implement than MAE.% for relatively fair comparison.

     \begin{figure}[t]
		\centering
		\includegraphics[width=0.465\textwidth]{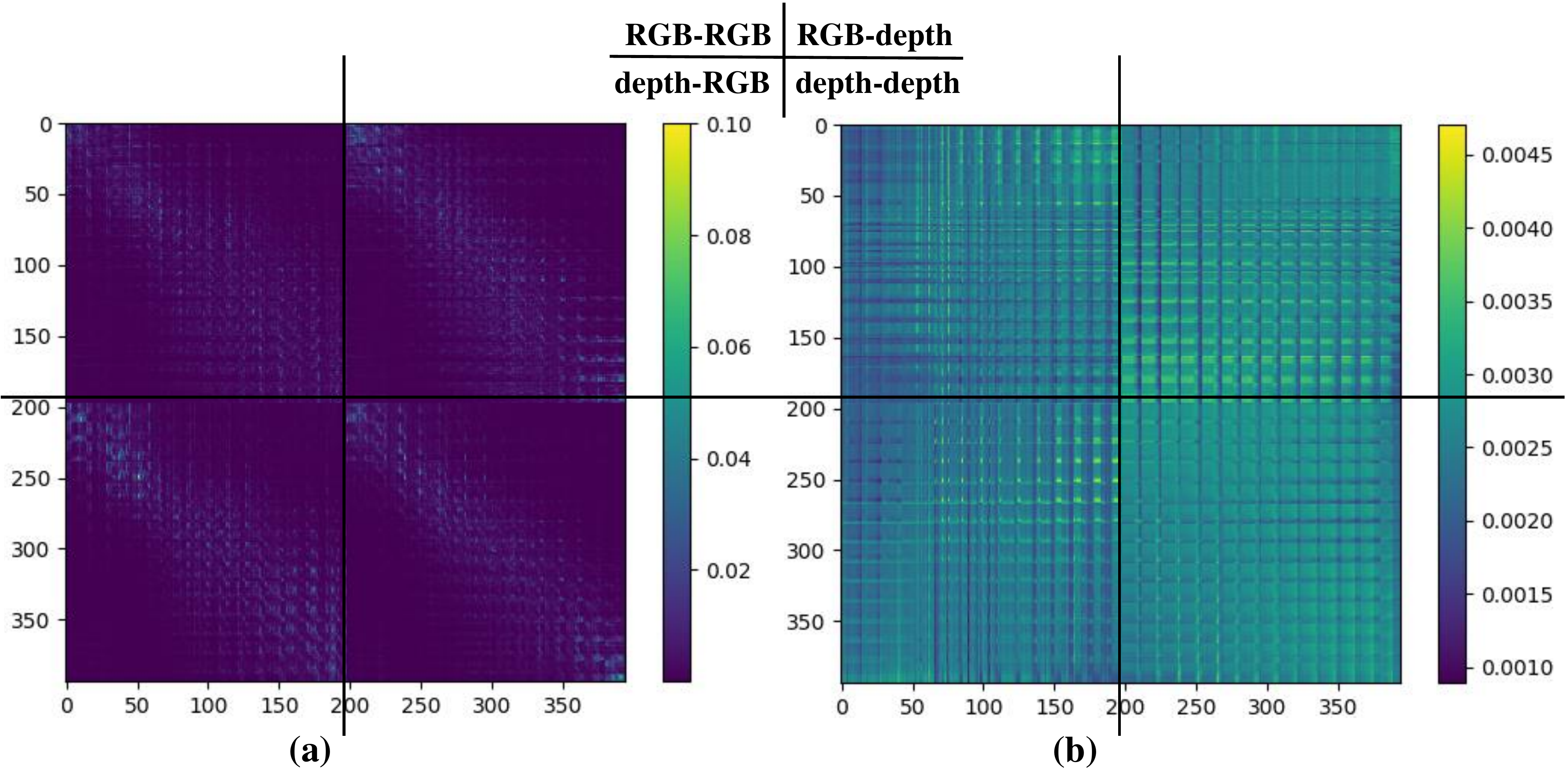} %  [width=0.9\columnwidth] Reduce the figure size so that it is slightly narrower than the column. Don't use precise values for figure width.This setup will avoid overfull boxes.
		\caption{Visualization of the attention maps of a representative attention head in encoder at the beginning of fine-tuning. (a) Transformer encoder with cross-modal patch-level contrastive pre-training. (b) Transformer encoder with random initialization. The four sub-squares from top to bottom and left to right in attention maps represent the attention between RGB and RGB, RGB and depth, depth and RGB, as well as depth and depth respectively. Due to the ability of inter-modal alignment and intra-modal local features extraction, the former focuses its attention around the main diagonal of the four sub-squares, while the latter is relatively uniform. Please refer to the supplementary material for more attention heads visualization.
  %It is recommended to view in magnified mode.
  }
		\label{visual}
	\end{figure}
	
	\subsection{Results on SUN RGB-D Dataset}
	We begin our experiments by studying the effectiveness of CoMAE for RGB-D scene recognition task on the SUN RGB-D dataset. In this subsection, unless otherwise specified, all pre-training and fine-tuning experiments are performed with ViT-B on SUN RGB-D training set.

\renewcommand{\dblfloatpagefraction}{.75}

     \subsubsection{Study on cross-modal patch-level contrastive pre-training.}

     \begin{table}[t]
		\centering
		\label{cross0}
		{
			\begin{tabular}{c|c|c}
				\toprule[2pt]
				\multirow{1}{*}{Method}   
				&\multicolumn{1}{c|}{${acc}_c$}
				&\multicolumn{1}{c}{${acc}_s$}\cr
				\hline
				From scratch         &40.5 &47.5 \\
				\hline
				Patch-level Contrast Pre-training         &\textbf{50.2} &\textbf{57.9} \\
				Instance-level Contrast Pre-training       &32.3 &40.8\\
				%\hline
				\bottomrule[2pt]
		    \end{tabular}
		}
		\caption{Ablation experiments of the cross-modal patch-level contrast pre-training.} 
		\label{cross}
   \end{table}

      In Tab. \ref{cross}, we verify the effectiveness of the proposed cross-modal patch-level contrastive pre-training. We train the same RGB-D transformer classification model from scratch as a baseline. Moreover, we implement a conventional cross-modal contrast framework which predicts which RGB goes with which depth in instance level. It can be clearly seen that the performance of our cross-modal patch-level contrastive pre-training significantly outperforms the other two methods. Conventional instance-level cross-modal contrastive learning fails since naive RGB-depth matching is a relatively easy pretext task and fails to capture high-level semantic information. In contrast, our cross-modal patch-level contrast can not only facilitate inter-modal alignment between RGB and depth, but also capture intra-modal local discriminative context features. We also visualize the attention maps of contrastive learning model in Fig.~\ref{visual}. Our patch-level contrastive objective can encourage the encoder to find the correspondence between RGB and depth patches.

      \begin{table}[t]
		\centering
		\label{fusion0}
		{
		\begin{tabular}{cc|c|c|c}
        %\hline
        \toprule[2pt]
        RGB & depth & Method & ${acc}_c$ & ${acc}_s$ \\ \hline
            \checkmark&       & RGB-MAE       & 42.9     &  52.7    \\ \hline
            &\checkmark       & depth-MAE        &  41.2    & 47.8     \\ \hline
            \checkmark&\checkmark      &AVG Fusion &   46.8     &  57.1          \\ \hline
            \checkmark&\checkmark     & MM-MAE &     \textbf{54.2}   &  \textbf{62.6}          \\ 
        \bottomrule[2pt]
        \end{tabular}
		    }
		\caption{Ablation experiments of the fusion paradigm of MM-MAE.} 
		\label{fusion}
	\end{table}

 \begin{table}[tbp]
		\centering
		\label{mae0}
		{\begin{tabular}{c|cc|c|c|c}\toprule[2pt]
                              & RGB & Depth & mask & ${acc}_c$ & ${acc}_s$ \\ \hline
              \multirow{3}{*}{scratch} & \checkmark   & \checkmark    &-      &40.5      &47.5      \\ \cline{2-6} 
                              & \checkmark   &       &-      &27.5      &34.8      \\ \cline{2-6} 
                              &     & \checkmark     &-      &38.6      &45.3      \\ \hline
             \multirow{4}{*}{MAE}     & \checkmark  & \checkmark     &mi      &\textbf{54.2}      &\textbf{62.6}      \\ \cline{2-6} 
                              & \checkmark   & \checkmark    &mc     &52.7      &61.6      \\ \cline{2-6} 
                              &\checkmark   &       &-      &42.9      &52.7      \\ \cline{2-6} 
                              &     &\checkmark    &-      &41.2      &47.8      \\ \bottomrule[2pt]
            \end{tabular}
			}
		\caption{ Ablation experiments of the multi-modal masked reconstruct pre-training. ’mi’, ’mc’ denotes mutual independency and mutual consistency respectively.}
		\label{mae}
	\end{table}

      \subsubsection{Study on multi-modal masked autoencoder.}
      
\begin{figure*}[ht]
	\centering
	\includegraphics[width=0.97\textwidth]{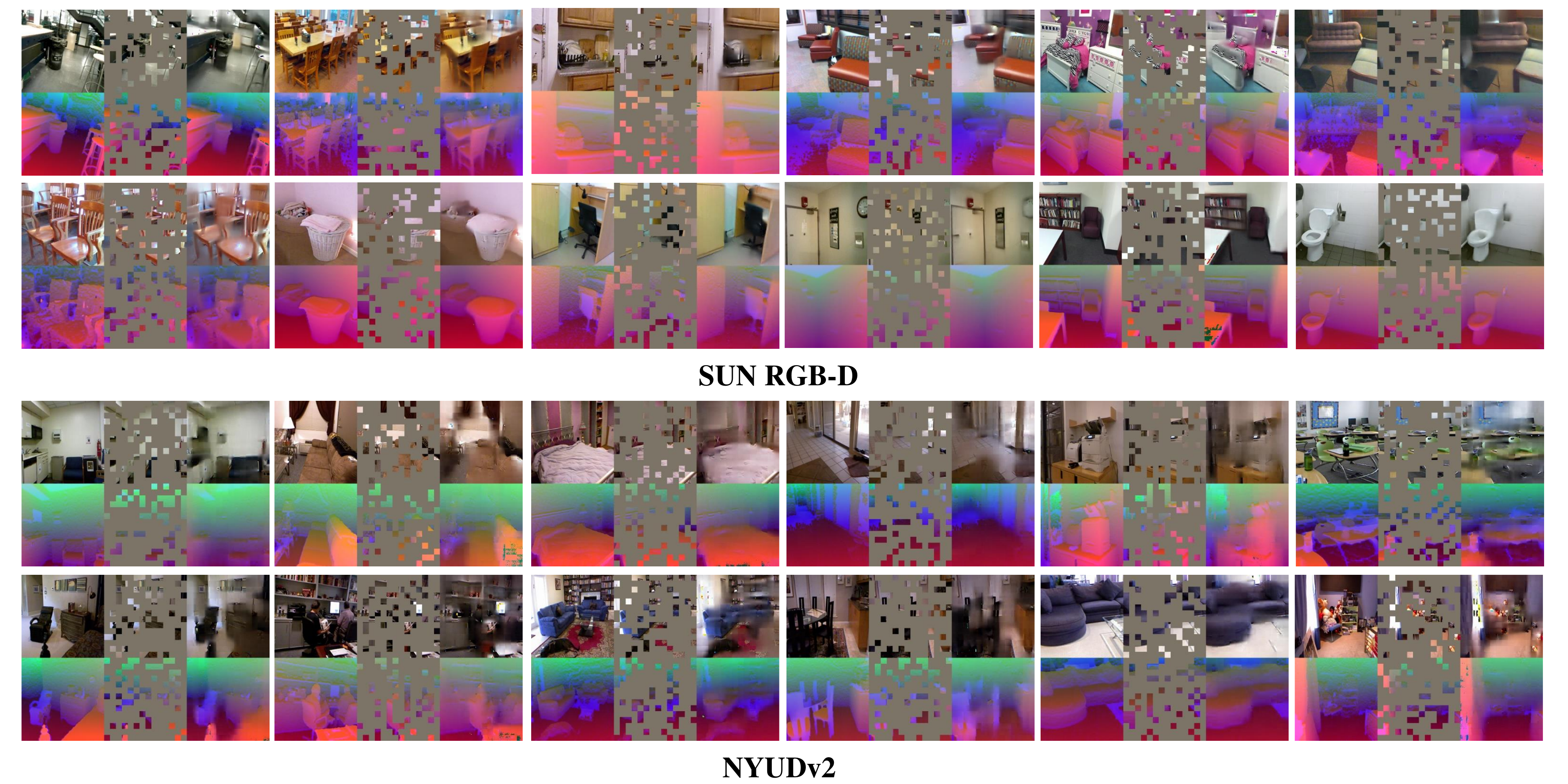} % Reduce the figure size so that it is slightly narrower than the column.
	\label{sun}
 	\caption{Some examples of reconstruction result on SUN RGB-D test set(Upon) and NYUDv2 test set(Down). For each triplet, we show the ground-truth (left), masked RGB-D (middle), and our MM-MAE reconstruction (right). We simply overlay the output with the visible patches to improve visual quality. The MM-MAE models are pre-trained using only their respective training set.}
	\label{vis2}
\end{figure*}

      In Tab. \ref{fusion}, we evaluate the effectiveness of the proposed multi-modal masked autoencoder. We compare the baseline method of training two separate MAEs for RGB and depth modalities, and then fine-tuning the corresponding encoder on the each modality. We see that the RGB MAE obtains the accuracy of 42.9\% and the depth modality performance is 41.2\%. We also report the fusion results of RGB and depth modality by taking an average of two MAE recognition results, and the performance is 46.8\%, which is worse than our CoMAE result. The superior performance of CoMAE demonstrates the effectiveness of jointly training a single encoder on two modalities and it can capture finer cross-modal structural information. We also provide the reconstruction results on SUN RGB-D and NYUDv2 in Fig. \ref{vis2}. 
      
      In addition, we explore two simple random masking strategies: mutual independency and mutual consistency of masked spatial location. As shown in Tab. \ref{mae}, the former is better although there exists the risk of indirect cross-modal information leakage in the masking and reconstruction pipeline. We suspect that the cross-modal information leakage could promote the multi-modal fusion, because the important and complementary information from cross-modal aligned patches is encouraged to capture in reconstruction task.

      \subsubsection{Study on the hybrid pre-training strategy.}
      After the specific ablations on the cross-modal patch-level contrastive pre-training and multi-modal masked autoencoder, we are ready to investigate the effectiveness of our proposed hybrid pre-training strategy of CoMAE. We compare with other two training alternatives: (1) joint training of transformer encoder with two objectives (CPC+MM-MAE), (2) the cascade training of transformer encoder in a reverse order (MM-MAE→CPC). The results are reported in Tab.~\ref{hybrid}. From results, we see that joint training could lead to worse performance than the single MM-MAE pre-training (50.6\% vs. 54.2\%). This is might be because that the patch-level cross-modal contrast discards positional embedding, which degrades its optimization consistency with multi-modal masked reconstruction and thus leads to inferior performance in downstream task. We also notice that the MM-MAE→CPC baseline obtains a worse performance than our CoMAE, which demonstrate the effectiveness of our curriculum learning design in our CoMAE, by learning easier task first and then training a harder task.
      
      %We perform ablation studies on different combination approaches of our cross-modal patch-level contrast and multi-modal masked reconstruction in Tab.\ref{hybrid}. The expriments demonstrate the advantages of our two-staged hybrid pre-training approach which performs contrast before reconstruction other than reverse order or conduct simultaneously. Nevertheless, the patch-level cross-modal contrast could capture the discriminative local context representations across RGB and depth modalities, which also greatly facilitates the 
     %further multi-modal masked reconstruction in a fusion manner owing to its inter-modal alignment prior.

     \begin{table}[t]
		\centering
		\label{hybrid0}
		{
			\begin{tabular}{c|c|c}
				\toprule[2pt]
				\multirow{1}{*}{Pre-training approach}   
				&\multicolumn{1}{c|}{${acc}_c$}
				&\multicolumn{1}{c}{${acc}_s$}\cr
				\hline
				CPC         &50.2 &57.9 \\
				MM-MAE         &54.2 &62.6 \\
				\hline
				CPC + MM-MAE        &50.6 &58.2\\
				MM-MAE → CPC        &51.4 &60.1\\
				CPC → MM-MAE        &\textbf{55.2} &\textbf{64.3}\\
				%\hline
				\bottomrule[2pt]
		    \end{tabular}
		    }
		\caption{Ablation experiments of the hybrid pre-training in a curriculum learning manner. 
  % CPC, MMR denotes cross-modal patch-level contrast and multi-modal masked reconstruction respectively. '+' denotes simultaneously execute between pretext tasks and '→' denotes providing initialization between pretext tasks.
  } 
		\label{hybrid}
	\end{table}

      \begin{table}[t]
		\centering
		\label{robust0}
		{
			 \begin{tabular}{cc|cc|cc|c|c}
			 	\toprule[2pt]
            \multicolumn{2}{c|}{Pre-train} & \multicolumn{2}{c|}{Fine-tune} & \multicolumn{2}{c|}{Test} & \multirow{2}{*}{${acc}_c$} & \multirow{2}{*}{${acc}_s$} \\
            R          & D          & R           & D          & R        & D        &                       & \multicolumn{1}{r}{}                      \\ \hline
            \checkmark            & \checkmark              & \checkmark            &\checkmark                & \checkmark          &\checkmark              &55.2                       &64.3                                           \\ \hline
            \checkmark            & \checkmark              & \checkmark            &                & \checkmark          &              &44.6                       &55.1                                           \\ \hline
            \checkmark            & \checkmark              &               & \checkmark              &            & \checkmark            &49.6                       &58.3                                           \\ \hline
            \checkmark            & \checkmark              & \checkmark             & \checkmark              & \checkmark          &              &38.1                       &43.8                                           \\ \hline
           \checkmark            &\checkmark              & \checkmark             & \checkmark              &            & \checkmark            & 39.2                      &47.0                                           \\ \hline 
           \checkmark            &\              & \checkmark             &               & \checkmark            &            &42.9                       &52.7                                           \\ \hline
                       &\checkmark              &             &\checkmark                &             &\checkmark            &41.2                       &47.8                                           \\ 
             \bottomrule[2pt]
             \end{tabular}
             }
		\caption{ Ablation experiments of the robustness of unimodal data. 'R’, ’D’ denotes RGB and depth respectively.}
		\label{robust}
	\end{table}

      \subsubsection{Study on the random modality dropping.}
      
      We propose a random modality dropping strategy with a probability of 0.5 as a very strong data augmentation in fine-tuning process. It is likely to degrade performance due to inconsistency between training and testing. However, our model benefits from this data augmentation in fine-tuning (55.2\% vs. 52.7\%). This result demonstrates that our model has great flexibility in handling modality inputs, and it can benefit from more unpaired unimodal data in fine-tuning phase.

      %这是和SUN RGB-D SOTA算法的比较 我们没有加载预训练模型 且没有使用额外数据（TrecgNet）它们小数据训不动大模型 8+
	\begin{table*}[ht]
		\centering
		\label{tab:freq6}
		{
			\begin{tabular}{c|c|c|c|c}
				\toprule[2pt]
				\multirow{1}{*}{Method}    &\multicolumn{1}{c|}{Backbone}   &\multicolumn{1}{c|}{Additional labeled training data}
				&\multicolumn{1}{c|}{${acc}_c$}
				&\multicolumn{1}{c}{${acc}_s$}\cr
				%&$F_\beta^{max} \uparrow$ &$mF_\beta \uparrow$  %&$S_m \uparrow$ &MAE$\downarrow$  %&$F_\beta^{max} \uparrow$ &$mF_\beta \uparrow$ %&$S_m \uparrow$ &MAE $\downarrow$\\
				\hline
				Multimodal fusion \cite{zhu2016discriminative}         &AlexNet &Places &41.5 &-\\
				Modality and component fusion \cite{wang2016modality}        &AlexNet &Places &48.1 &-\\
				RGB-D-CNN (wSVM) \cite{song2017depth} &AlexNet &Places &52.4 &-\\
				{RGB-D-OB (wSVM)}$^{\mbox{\scriptsize*}}$ \cite{song2018learning}   &AlexNet &Places &53.8 &- \\
				{G-L-SOOR}$^{\mbox{\scriptsize*}}$ \cite{song2019image}   &AlexNet &Places &55.5 &- \\
				DF$^{\mbox{\scriptsize2}}$Net \cite{li2018df} &AlexNet &Places &54.6 &- \\
				ACM \cite{yuan2019acm}   &AlexNet &Places &55.1 &- \\
				MSN \cite{xiong2020msn}  &AlexNet &Places &56.2 &- \\
				TRecgNet  \cite{du2019translate}  &ResNet18 &Places, 5K unlabeled, gen &56.7 &- \\
				%TRecgNet (jn) \cite{du2021cross}  &ResNet101 &ImageNet, 5K unlabeled, generated data &59.8 &- \\
				ASK \cite{xiong2021ask} &ResNet18 &Places & {\bf 57.3} &- \\
				CNN-randRNN \cite{caglayan2022cnns}  &ResNet101 &ImageNet &- &60.7 \\
				Omnivore \cite{girdhar2022omnivore} &Swin-B &ImageNet-22K, Kinetics-400 &- & {\bf 67.2} \\
				Omnivore \cite{girdhar2022omnivore} &Swin-L &ImageNet-22K, Kinetics-400 &- &67.1 \\
				\hline
				\makecell[c]{Ours (w/o pre-training)} &ViT-B	&None	&40.5 &47.5 \\
				\makecell[c]{Ours (CoMAE pre-training)} &ViT-S	&None	&52.6 &62.1 \\
				\makecell[c]{Ours (CoMAE pre-training)} &ViT-B	&None	&\textbf{55.2} &\textbf{64.3} \\
				\makecell[c]{Ours (Image MAE pre-training)} &ViT-B	&ImageNet-1K (unlabeled)	&55.2 &63.6 \\
				\bottomrule[2pt]
		\end{tabular}
		}
		\caption{Comparison with state-of-the-art methods on SUN RGB-D. * denotes using object detection to obtain object annotations. 'gen' denotes using generated data for the training data sampling enhancement.} 
		\label{sotasun}
	\end{table*}

		%可能得采取一下措施，用视频数据集或者加载SUN RGB-D的
	\begin{table*}[!h]
		\centering
		\label{tab:freq7}
		{
			\begin{tabular}{c|c|c|c|c}
				\toprule[2pt]
				\multirow{1}{*}{Method}    &\multicolumn{1}{c|}{Backbone}   &\multicolumn{1}{c|}{Additional labeled training data}
				&\multicolumn{1}{c|}{${acc}_c$}
				&\multicolumn{1}{c}{${acc}_s$}\cr
				%&$F_\beta^{max} \uparrow$ &$mF_\beta \uparrow$  %&$S_m \uparrow$ &MAE$\downarrow$  %&$F_\beta^{max} \uparrow$ &$mF_\beta \uparrow$ %&$S_m \uparrow$ &MAE $\downarrow$\\
				\hline
				Modality and component fusion \cite{wang2016modality}        &AlexNet &Places, SUN RGB-D &63.9 &-\\
				RGB-D-CNN \cite{song2017depth} &AlexNet &Places &65.8 &-\\
				{RGB-D-OB (wSVM)}$^{\mbox{\scriptsize*}}$ \cite{song2018learning}   &AlexNet &Places &67.5 &- \\
				{G-L-SOOR}$^{\mbox{\scriptsize*}}$ \cite{song2019image}   &AlexNet &Places, SUN RGB-D &67.4 &- \\
				DF$^{\mbox{\scriptsize2}}$Net \cite{li2018df} &AlexNet &Places &65.4 &- \\
				ACM  \cite{yuan2019acm}   &AlexNet &Places &67.4 &- \\
				MSN \cite{xiong2020msn}  &AlexNet &Places &68.1 &- \\
				TRecgNet  \cite{du2019translate}  &ResNet18 &Places, SUN, 5K unlabeled, gen &69.2 &- \\
				%TRecgNet (jn) \cite{du2021cross}   &ResNet18 &Places, 5K unlabeled, generated data &71.8 &- \\
				ASK \cite{xiong2021ask} &ResNet18 &Places, SUN & {\bf 69.3} &- \\
				Omnivore \cite{girdhar2022omnivore} &Swin-B &ImageNet-22K, Kinetics-400, SUN &- &79.4 \\
				Omnivore \cite{girdhar2022omnivore} &Swin-L &ImageNet-22K, Kinetics-400, SUN &- & {\bf 79.8} \\
				\hline
				\makecell[c]{Ours (w/o pre-training)} &ViT-B	&None	&40.7 &49.8 \\
				\makecell[c]{Ours (CoMAE pre-training)} &ViT-S	&None	&61.4 &66.8 \\
				\makecell[c]{Ours (CoMAE pre-training)} &ViT-B	&None	&62.7 &66.5 \\
				\makecell[c]{Ours (SUN RGB-D pre-training)} &ViT-S	&SUN RGB-D	&68.6 &71.4 \\
				\makecell[c]{Ours (SUN RGB-D pre-training)} &ViT-B	&SUN RGB-D	&\textbf{70.7} &\textbf{76.3} \\
				%\makecell[c]{Ours (Image MAE pre-training)} &ViT-B	&ImageNet-1K (unlabeled)	&46.7 &54.0 \\
				\bottomrule[2pt]
		\end{tabular}
		}
		\caption{Comparison with state-of-the-art methods on NYUDv2. * denotes using object detection to obtain object annotations. 'gen' denotes using generated data for the training data sampling enhancement.} 
		\label{sotanyu}
	\end{table*}
	% * means use object detection model to obtain object annotations. cf refers to the results from the conference version. jn refers to the results from the journal version.

 \renewcommand{\dblfloatpagefraction}{.9}

 %      \begin{table}[htbp]
	% 	\centering
	% 	\label{tab:freq4}
	% 	{
	% 		\begin{tabular}{c|c|c}
	% 			\toprule[2pt]
	% 			\multirow{1}{*}{strong data augmentation}   
	% 			&\multicolumn{1}{c|}{${acc}_c$}
	% 			&\multicolumn{1}{c}{${acc}_s$}\cr
	% 			\hline
	% 			w/o          &52.7 &62.4 \\
	% 			randomly modality droping         &\textbf{55.2} &\textbf{64.3} \\
	% 			%\hline
	% 			\bottomrule[2pt]
	% 	    \end{tabular}
	% 	    }
	% 	\caption{Ablation experiments of the data augmentation of randomly modality droping in fine-tuning.} 
	% 	\label{finetune}
	% \end{table}

      \subsubsection{Study on the robustness of unimodal data.}
      
    It is difficult to collect paired multimodal data from the real world, due to the unavailability of a sensor. Therefore multimodal models are expected to show robustness against unimodal data.
    As shown in Tab.~\ref{robust}, we evaluate our CoMAE's robustness of unimodal data in fine-tuning and test phases. Different from some existing methods, our CoMAE is very flexible with its training and testing setting with multimodal or unimodal data. Surprisingly, our pre-trained multi-modal model significantly outperforms RGB or depth single-modal model pre-trained by image masked autoencoder under the fine-tuning and testing with only RGB or depth modality. These results demonstrate that our hybrid multimodal pre-training can significantly promote unimodal learning, and we suspect that our CoMAE pre-training can enforce model implicitly obtain some complementary modal information when only using unimodal data to inference.

   \subsubsection{Comparison with the state-of-the-art methods.}
      
      We report the performance of our CoMAE on SUN RGB-D test set and compare with state-of-the-art methods in Tab. \ref{sotasun}. Although these methods utilize supervised pre-trained models or many additional labeled data, our CoMAE can still achieve competitive results without utilizing any extra data or pre-trained models, demonstrating the effectiveness of our hybrid pre-training approach on relatively small-scale datasets. This property is quite important in real applications as sometimes it is really hard to collect lots of samples. In addition, our CoMAE instantiated with ViT-B also achieves competitive and even better results of the same ViT-B model pre-trained by ImageMAE \cite{he2022masked}, this reveals the data efficiency advantage of our CoMAE pre-training.

	\subsection{Results on NYUDv2 Dataset}
	We also evaluate the performance of our CoMAE on the NYUDv2 test set and compare with other methods in Tab.~\ref{sotanyu}. NYUDv2 is very small and the training data is seriously imbalanced, and consequently some works load the weights pre-trained on SUN RGB-D dataset for model initialization on NYUDv2 dataset. Our CoMAE achieves much better result than training from scratch on NYUDv2 when only using its training set for pre-training, but it is not excellent enough. We conjecture that NYUDv2 train set is too small to fully unleash the pre-training power of CoMAE. In particular, it lacks enough labeled data for explicit semantic supervision during fine-tuning, which is very critical for masked image modeling self-supervised learning paradigms. Therefore, following the previous methods, we transfer the pre-trained model from SUN RGB-D training set and fine-tune on NYUDv2 training set, achieving excellent results, which confirms our conjecture and also verifies the generalization ability of our CoMAE.

	\section{Conclusions and Future Works}
	 In this work, we have proposed CoMAE, a data-efficient and single-model self-supervised hybrid pre-training framework for RGB and depth representation learning. Our CoMAE presents a curriculum learning strategy to unify two types of self-supervised learning methods, which is composed of two critical pretext tasks of cross-modal patch-level contrast and multi-modal masked reconstruction. In addition, our single model design without requirement of fusion module is very flexible and robust to generalize to unimodal data, and has significant practical value for scenarios with limited modality available. Extensive experiments on SUN RGB-D and NYUDv2 datasets demonstrate the effectiveness of our CoMAE on multi-modal representation learning, in particular for small-scale datasets. Our CoMAE simply pre-trained on small-scale RGB-D datasets obtains very competitive performance to those models with large-scale pre-training.
	 
	 As the first method for self-supervised pre-training on small-scale RGB-D datasets, we hope it can be taken as a strong baseline and facilitate further research along this direction, in particular for research groups with limited computational resources. Finally, we believe that our CoMAE has a large space for exploration in negative examples mining and masking strategies of spatial location relationships between modalities. In addition, transferring our pre-trained ViT for downstream dense prediction tasks such as RGB-D semantic segmentation is also possible future work.

%     \begin{table*}[htbp]
%     \centering
%     \begin{subtable}[t]{0.495\linewidth}
%     \centering
%       \scalebox{0.65}{
% 			\begin{tabular}{c|c|c}
% 				\toprule[2pt]
% 				\multirow{1}{*}{Method}   
% 				&\multicolumn{1}{c|}{${acc}_c$}
% 				&\multicolumn{1}{c}{${acc}_s$}\cr
% 				\hline
% 				from scratch         &40.5 &47.5 \\
% 				\hline
% 				patch-level contrast pre-training         &\textbf{50.2} &\textbf{57.9} \\
% 				instance-level contrast pre-training       &32.3 &40.8\\
% 				%\hline
% 				\bottomrule[2pt]
% 		    \end{tabular}
% 		    }
%         \caption{Ablation experiments of the cross-modal patch-level contrast.}
%     \end{subtable}
%     \begin{subtable}[t]{0.495\linewidth}
%     \centering
%         \scalebox{0.65}{
% 			\begin{tabular}{c|c|c}
% 				\toprule[2pt]
% 				\multirow{1}{*}{strong data augmentation}   
% 				&\multicolumn{1}{c|}{${acc}_c$}
% 				&\multicolumn{1}{c}{${acc}_s$}\cr
% 				\hline
% 				w/o          &52.7 &62.4 \\
% 				randomly modality droping         &\textbf{55.2} &\textbf{64.3} \\
% 				%\hline
% 				\bottomrule[2pt]
% 		    \end{tabular}
% 		    }
%         \caption{Ablation experiments of the data augmentation of randomly modality droping in fine-tuning}
%     \end{subtable}
%     \caption{Caption here}
%     \label{tab:array}
% \end{table*}

\section{Acknowledgements}
This work is supported by National Natural Science Foundation of China (No. 62076119, No. 61921006), the Fundamental Research Funds for the Central Universities (No. 020214380091), and Collaborative Innovation Center of Novel Software Technology and Industrialization.

%还可以写一些
% This work is supported by National Natural Science Foundation of China (No. 62076119, No. 61921006), the Fundamental Research Funds for the Central Universities (No. 020214380091), and Collaborative Innovation Center of Novel Software Technology and Industrialization.

\bibliography{aaai23}

\end{document}